\documentclass{article}

\PassOptionsToPackage{numbers, compress}{natbib}
\usepackage[preprint]{neurips_2024}

\usepackage[utf8]{inputenc} 
\usepackage[T1]{fontenc}    
\usepackage{hyperref}       
\usepackage{url}            
\usepackage{booktabs}       
\usepackage{amsfonts}       
\usepackage{nicefrac}       
\usepackage{microtype}      
\usepackage{xcolor} 
\usepackage{amsmath}
\usepackage{multirow} 
\usepackage{subcaption} 
\usepackage{graphicx} 

\usepackage{caption}

\title{Exploring Molecule Generation \\ Using Latent Space Graph Diffusion}

\author{%
   Prashanth Pombala\\
  Saarland Informatics Campus \\
  Saarland University \\
  66123 Saarbrücken\\
  Germany \\
  \texttt{prpo00001@uni-saarland.de} \\
   \And
   Gerrit Großmann \\
   DFKI \\
   67663 Kaiserslautern \\
   Germany \\
   \texttt{gerrit.grossmann@dfki.de} \\
   \AND
   Verena Wolf \\
   Saarland Informatics Campus \\
   DFKI \\
   66123 Saarbrücken \\
   Germany \\
   \texttt{verena.wolf@dfki.de} \\
}

\begin{document}

\maketitle

\begin{abstract}
    Generating molecular graphs is a challenging task due to their discrete nature and the competitive objectives involved. Diffusion models have emerged as SOTA approaches in data generation across various modalities. For molecular graphs, graph neural networks (GNNs) as a diffusion backbone have achieved impressive results. Latent space diffusion—--where diffusion occurs in a low-dimensional space via an autoencoder—--has demonstrated computational efficiency. However, the literature on latent space diffusion for molecular graphs is scarce, and no commonly accepted best practices exist. In this work, we explore different approaches and hyperparameters, contrasting generative flow models (\emph{denoising diffusion}, \emph{flow matching}, \emph{heat dissipation}) and architectures (\emph{GNNs} and \emph{E(3)-equivariant GNNs}). Our experiments reveal a high sensitivity to the choice of approach and design decisions. Code is made available at {\hyperlink{https://github.com/Prashanth-Pombala/Molecule-Generation-using-Latent-Space-Graph-Diffusion}{\scriptsize \texttt{github.com/Prashanth-Pombala/Molecule-Generation-using-Latent-Space-Graph-Diffusion}}}.  

\end{abstract}

\section{Introduction}

\begin{figure}[htbp]
  \centering
  \includegraphics[width=0.99 \textwidth]{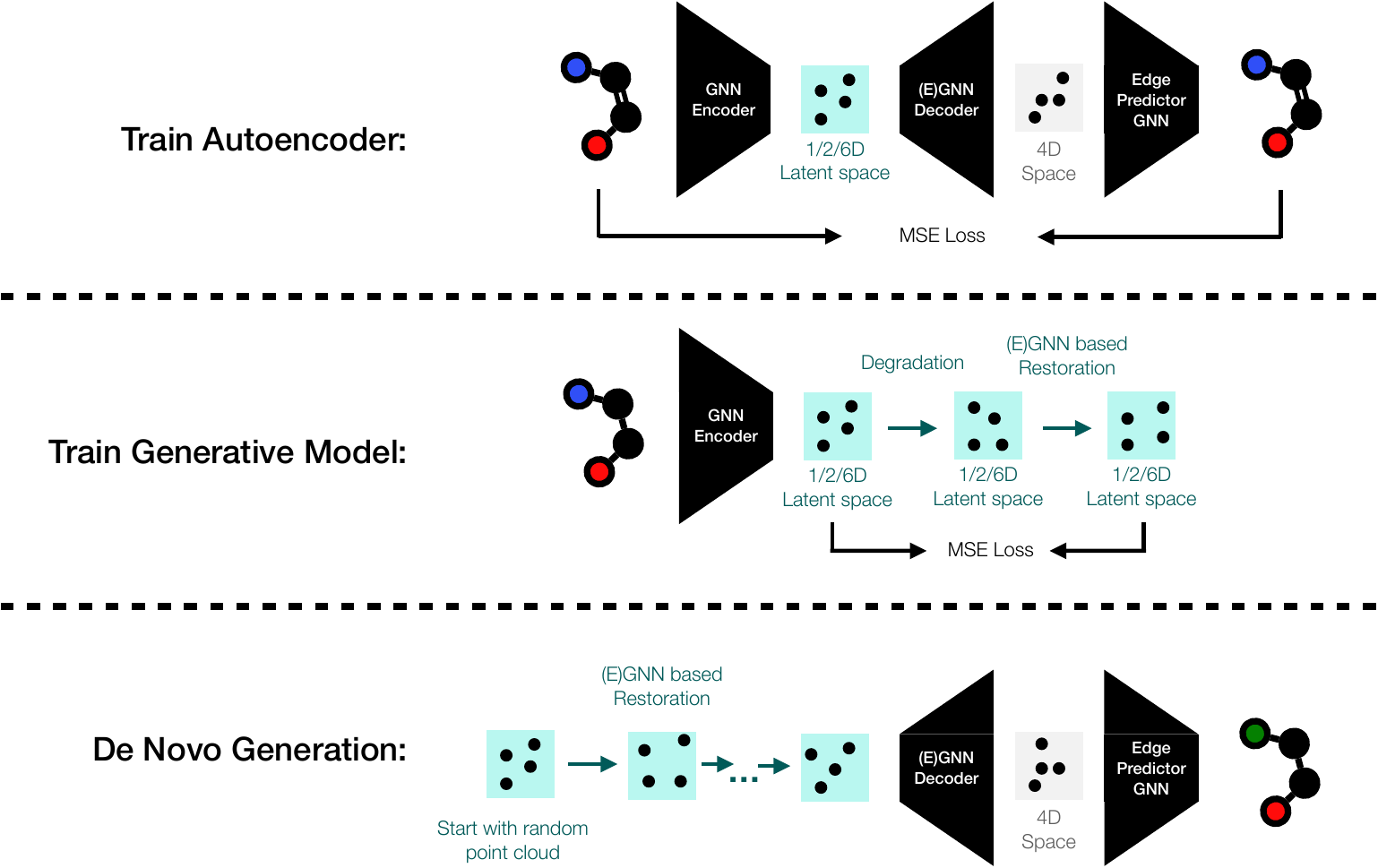}
  \caption{Schematic overview of our training process. (1) An autoencoder is trained to embed atoms into a low-dimensional latent space. Molecule reconstruction involves two steps: predicting a feature vector for each atom, and predicting molecular bonds for each pair of atoms. (E)GNN means that either a vanilla GNN or an EGNN is used.  (2) Information in the latent space is removed using a degradation operator (e.g., adding noise), and an (E)GNN is trained to restore the original data. (3) For \emph{de novo} molecule generation, we start with random noise (or the endpoint of the degradation process) and iteratively add information using the (E)GNN until we obtain a final point cloud, which is decoded into a molecule.}
  \label{fig:mainfig}
\end{figure}

Drug discovery involves identifying molecules capable of interacting with specific molecular targets in the body to treat diseases. Generative AI methods are increasingly employed for molecule generation. Recently, the rapid growth in image diffusion models has inspired various approaches for molecule generation~\cite{xu2023geometric,hoogeboom2022equivariant}. Each method presents unique strengths and challenges, contributing to a rapidly evolving field. However, the growing diversity of techniques and parameters has made it increasingly challenging and time-consuming to identify the most effective approach for a given application.

We propose that key factors significantly influencing molecule generation include the choice between latent space and traditional graph diffusion methods, the dimensionality of the latent space, the selection of various diffusion processes, and the decision between using \emph{Graph Neural Networks} (GNNs)~\cite{scarselli2009graph} or \emph{E(3)-Equivariant Graph Neural Networks} (EGNNs)~\cite{satorras2021en}, which maintain equivariance for rotations, translations, and reflections in the Euclidean space. This means that we interpret the input data effectively as a point cloud and process it using pair-wise distances. Additionally, we explore alternative diffusion processes, such as heat equation-based diffusion~\cite{rissanen2022generative} and flow matching~\cite{lipman2022flow}.  We also propose a novel architecture that separates bond type prediction from edge prediction, where an edge-type prediction model predicts the edge type from the generated molecular graph resulting in more accurate molecule generation.  Our goal is to identify the most advantageous strategies for enhancing molecular generation, providing insights to guide future research and applications in the field.

\section{Related Work}
\paragraph{Diffusion.}
Denoising Diffusion Probabilistic Models \cite{ho2020denoising}, represent a novel approach to generative modeling that employs a diffusion process that gradually degrades the data and then learns to reverse this degradation to generate new data samples. Beyond standard Gaussian diffusion~\cite{rissanen2022generative} implements generative modeling using heat equation-based diffusion for images. The work of~\cite{lipman2022flow} proposes flow matching for generative modeling of point clouds, showing promising results. This study explores the potential of these methods for graph generation tasks and implements them. Latent diffusion models perform diffusion in the latent space~\cite{rombach2022high} and advance denoising diffusion models by showcasing improved performance across various aspects of image generation.
\paragraph{Molecule Generation.}
Molecule generation using diffusion models has gained significant attention in recent years, leading to a diverse array of approaches and methodologies. The seminal paper by~\cite{ho2020denoising} inspires adaptations of denoising diffusion models to various graph generation tasks. Recent work, such as~\cite{xu2023geometric}, adapts latent diffusion models for molecule generation. 
\paragraph{Graph Autoencoding.}
In~\cite{kipf2016variational}, the authors discuss graph autoencoders, which encode graph structures into a latent space representation and then decode them to reconstruct the original graph. The paper~\cite{corso2020principal} discusses graph neural networks based on principal neighbourhood aggregation,  which can be used to obtain improved accuracy.

\section{Method}
This section provides an overview of the method used in this work. Appendix~\ref{appendix} provides detailed information.

\paragraph{Molecule Representation.}
\label{subsec:Molecule Representation}

We represent each molecule as a molecular graph, where nodes correspond to atoms and edges indicate the presence of a covalent bond between two atoms. We label each edge to indicate the bond type (single, double, triple, or ring). Additionally, we represent each node (atom) with a 4-dimensional vector encoding its element (C, O, N, F) using a one-hot scheme. We let $n$ denote the number of nodes. Following standard practice, we do not include hydrogen atoms.

\paragraph{Autoencoder.}
\label{subsec:autoencoder}

We use autoencoders to learn a latent representation of each molecule. The latent representation of a molecule is a point cloud composed of the latent representations of its individual atoms. The encoder learns to map each atom to its latent representation, given the entire molecule as context. The decoder then learns to reconstruct the molecular graph from this point cloud using either a classical GNN or an EGNN architecture.
The encoder and decoder both use the PNA architecture~\cite{corso2020principal}.

\paragraph{Encoder.}

The encoder generates a $z$-dimensional latent space embedding for the input molecules (i.e., a position when interpreted as a point cloud in latent space; $z$ ranges over $1,2,6$) for each atom using a 2-layer PNA architecture. The input molecules are represented as described above.

\paragraph{GNN-Based Decoder.}
\label{para:GNN}
The GNN-based decoder consists of three steps. (1) We convert the point cloud of latent embeddings into a complete graph, where the node features depend on the atom embeddings (i.e., the locations). (2) We then feed this graph into a 2-layer PNA decoder, which translates the embeddings into 4-dimensional atom representations. (3) Next, we feed all pairs of atom embeddings into an edge prediction layer (a 2-layer MLP), which predicts the presence of an edge between them based on a threshold. Finally, we predict the bond type (single, double, triple, ring) by giving as input the decoded molecular graph to a 2-layer graph convolutional network model followed by a 2-layer MLP. Note that the bond-type predictor deterministically selects a single possible bond type, given the molecular graph and following valency rules.

\paragraph{EGNN-Based Decoder.}
The EGNN-based decoder consists of three steps: (1) We construct a graph by introducing dummy nodes to represent edges between each pair of point clouds in the latent space, with features of the dummy nodes based on the pairwise L2 distances of the point cloud latent embeddings. (2) We feed this graph into a 2-layer PNA decoder, using the output features of the dummy nodes to infer the presence or absence of an edge between two nodes based on a threshold. (3) Finally, we predict the bond type using the same method as in the GNN-based decoder.

\subsection{Flavors of Diffusion}
\label{subsec:diffusion}
This section outlines the diffusion processes used in our experiments, following the notation from~\cite{bansal2023cold}. We refer to the information removal process as \emph{degradation} rather than noising (or \emph{forward} process) and use \emph{restoration} to describe the information-adding process via a neural network backbone, rather than denoising (or \emph{reverse} process). Figure~\ref{fig:mainfig} provides an overview of the training and generation procedures.

The first process we applied is standard \textbf{Gaussian-based diffusion} \cite{ho2020denoising}. The degradation process iteratively adds Gaussian noise to the data, gradually pushing it toward an isotropic Gaussian centered around zero. In the restoration process, a GNN processes the degraded embeddings and their corresponding timestep as input to predict the restored embeddings. Using these predictions, the model refines the data by progressively adjusting it toward the restored state. In the generation process, the model begins with fully noised embeddings sampled from a Gaussian distribution and restores them step-by-step in 50 timesteps until it achieves the fully restored representation.

For EGNN diffusion, we used the restoration model defined by \cite{xu2023geometric}. Starting from the fully degraded embeddings, the restoration model takes in the degraded embeddings, computes the pairwise distance between the embeddings, and uses this information to predict the restored embeddings. Using these predictions, the model refines the data by progressively moving it toward the restored state.

The second process we used is \textbf{heat dissipation} \cite{rissanen2022generative}, which simulates heat dissipation and then reverses it to restore the original data. This method consists of two stages: a degradation process that blurs information and a restoration process that reverses this blurring to recover the data. The degradation process blurs embeddings using the heat equation. First, we transform the embeddings into the frequency domain using the Discrete Cosine Transform. In this domain, we attenuate high-frequency components by applying a decaying exponential function, simulating the effect of heat dissipation. We then transform the blurred embeddings back into the spatial domain using the Inverse Discrete Cosine Transform. In the restoration process, a GNN reverses the heat equation to restore the embeddings. At each timestep, the model takes the blurred embeddings and predicts the difference needed to restore them to a less blurred state, similar to the embeddings from the previous timestep. In the generation process, we start with blurred embeddings derived from the heat dissipation mechanism. Using the restoration model, we iteratively refine the embeddings. At each step, we add Gaussian noise. This iterative process continues until the embeddings are fully restored.

The third approach, \textbf{flow matching}~\cite{lipman2022flow}, includes a training phase and a generation phase. In the training phase, we construct a vector field to represent the instantaneous velocity required to transform points from a simple source distribution to a complex target distribution. We construct the vector field by interpolating between the source and target embeddings to estimate the sample's position at any given time. The vector field at each point captures the required velocity to align the interpolated position with the target embedding. The GNN model processes node embeddings over a complete graph structure to predict the vector field, representing the instantaneous velocity of the transformation. The training objective reduces the discrepancy between the GNN-predicted vector field and the defined vector field. In the generation phase, the model maps samples from a standard normal distribution and uses an ODE solver to transform them into the target distribution by integrating the learned vector field.

\section{Results}
All models were trained on the QM9 dataset containing 133,885 
molecules with up to nine heavy (i.e., non-hydrogen) atoms.

During the sampling process, we generated molecules and calculated their validity, uniqueness, and novelty. Validity assesses whether the generated molecules adhere to fundamental chemical rules, such as valency and bond structure. Uniqueness measures the diversity within the set of generated valid molecules, defined as the proportion of distinct molecules relative to the total number of generated molecules. Novelty evaluates the number of generated molecules that are new, meaning they do not appear in the training set.

The following sections detail the experiments and their results. We randomly selected the number of molecules generated within the range of 100 to 500 and averaged the results for consistency.

\paragraph{Experiments.} 
In Experiment 1, we applied GNN-based latent space Gaussian graph diffusion, with latent space dimensions of two and six. The model achieved strong validity with moderate training demands, though it exhibited lower uniqueness. In Experiment 2, we applied EGNN-based latent space Gaussian graph diffusion, with latent space dimensions of two and six. The model achieved higher uniqueness but lower validity, requiring significantly more computational resources. In Experiment 3, we used GNN-based Gaussian graph diffusion in the input space, with embedding dimensions of two and six. This model was the most resource-intensive. In Experiment 4, we applied heat equation-based diffusion in a 1D embedding space. This method produced the highest novelty at a minimal computational cost but had the lowest uniqueness. In Experiment 5, we used GNN-based latent space flow matching with latent space dimensions of two and six. This model demonstrated balanced performance across all metrics.

\begin{figure}[h!]
    \centering
    \includegraphics[width=0.8\textwidth]{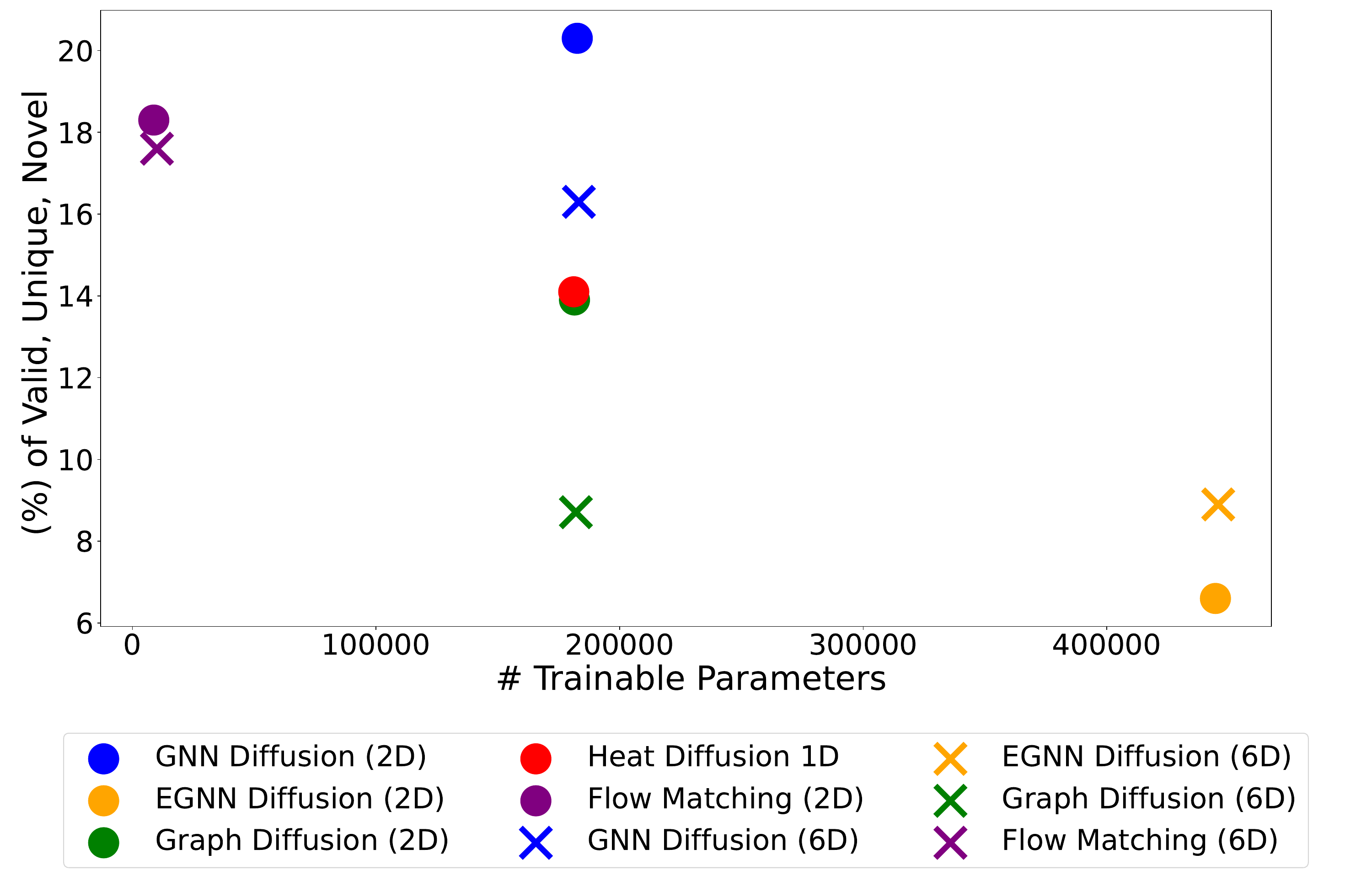}
    \caption{This scatter plot shows the relationship between the number of trainable parameters and the percentage of valid, unique, and novel molecules generated. The graph highlights that different model configurations show varying performance and model performance is not always dependent on the number of parameters.}
    \label{fig:trainable_parameters}
\end{figure}

Figure \ref{fig:trainable_parameters} shows a scatter plot illustrating the relationship between the number of trainable parameters and the total percentage of valid, unique, and novel molecules.

\section{Discussion}

We applied the GNN-based latent space Gaussian graph diffusion model across 2D and 6D configurations. In the 2D configuration, the model achieved high validity, balanced novelty, and moderate uniqueness, while training times per epoch stayed relatively efficient. At 6D, validity decreased while uniqueness improved, with novelty remaining stable. The minimal increase in training time at this dimensionality suggests that higher dimensions introduced a manageable computational cost.

The EGNN-based latent space Gaussian graph diffusion model, using the EGNN autoencoder described in Section~\ref{subsec:autoencoder} and the EGNN diffusion model described in Section~\ref{subsec:diffusion}, achieved high uniqueness and novelty but lower validity in the 2D configuration. This model required notably more training time than the GNN-based model, making it more computationally intensive. At 6D, uniqueness and novelty remained high, but validity declined further, and training times increased considerably. This result shows that EGNN-based diffusion generates unique structures but requires significant computational resources, especially at higher dimensions. Our adaptation of the EGNN to operate on edge indices instead of coordinates likely reduced the validity of the generated molecules compared to GNN-based methods.

The GNN-based Gaussian graph diffusion model in input space achieved high uniqueness and novelty but lower validity in the 2D configuration, with a substantially longer training time per epoch. This observation indicates that input-space diffusion is highly computationally intensive. In the 6D configuration, validity improved slightly while uniqueness and novelty declined, with training demands remaining high, underscoring this method’s intensive computational requirements.

The 1D heat equation-based diffusion model achieved high validity but had lower uniqueness and high novelty. Training time for this model stayed relatively efficient compared to other methods, making it computationally light. This approach may be advantageous in scenarios prioritizing validity over diversity, as it generates highly valid but more repetitive molecular structures at a low computational cost. Heat equation-based models exhibit high validity because they start with real data, blur the data in the degradation process, and deblur it in the restoration process to generate new molecules. Starting with a valid data point instead of random embeddings helps in generating chemically valid outputs and conforming to learned structures. However, this approach limits the model's ability to produce highly unique or novel outputs.

The GNN-based latent space flow matching model demonstrated balanced performance across all metrics in the 2D configuration, with manageable training times. In the 6D configuration, validity decreased slightly while uniqueness and novelty stayed high, with only a moderate increase in computational demand. This method achieved a strong balance of structural accuracy and diversity with moderate computational requirements.

Increasing latent space dimensions typically reduced validity because added complexity diluted the model’s ability to learn. Higher-dimensional spaces also correlated with increased training time, particularly in EGNN and input-space GNN-based models, highlighting greater computational inefficiencies in these configurations. These findings suggest that it is feasible to adequately encode the topology of small molecules within two dimensions. Another feature observed with GNN models was that increasing the size of the generated molecules improved uniqueness while maintaining validity.

The observed trade-offs between validity, uniqueness, and novelty emphasize the importance of selecting a model based on specific application goals, such as prioritizing structural accuracy or molecular diversity in generated outputs.

The work Geometric Latent Diffusion Models for 3D Molecule Generation (GeoLDM)~\cite{xu2023geometric} remains a strong benchmark for assessing diffusion models, given its robust performance metrics. Specifically, GeoLDM reports a validity of 93.8\% and a combined validity and uniqueness score of 92.7\%, demonstrating its ability to generate reliable and unique molecular structures. In comparison, our models did not achieve the level of combined validity and uniqueness shown by GeoLDM. GeoLDM uses atom coordinates as inputs, which provide the model with direct spatial information on the molecular structure. In contrast, we used the edge index as input, which focuses on the connectivity properties of the graph rather than precise spatial coordinates. This choice influences how the models interpret and generate molecular structures. GeoLDM uses a variational autoencoder to encode data into a latent space with built-in probabilistic variability. Our approach, on the other hand, relies on a standard autoencoder, which lacks the probabilistic component and may limit the diversity of generated outputs. These distinctions in data representation and autoencoder type lead to GeoLDM's higher uniqueness and validity scores.

\section{Conclusion and Future Work}

We identify several open directions for enhancing the molecule generation model and evaluation framework. Expanding the metrics beyond validity, uniqueness, and novelty allows for a more comprehensive evaluation of generation quality. While these metrics capture general properties, they fail to fully account for factors such as molecular stability, synthetic feasibility, or specific functional group accuracy, which play a critical role in practical applications in drug discovery and material science. For example, a molecule may be valid and novel but inherently unstable or difficult to synthesize, limiting its real-world utility.

Additionally, future work may explore alternative diffusion methods and include them in comparative studies. Future work may also design deeper autoencoder models that maintain low variance in embeddings. Another potential direction involves integrating the edge-type prediction process directly within the decoder rather than using a dedicated model for edge-type prediction. Future work may also increase the dimensionality of our 1D heat equation-based model to assess how this change impacts the results. Appendix~\ref{appendix} provides additional details about the experiments.

\bibliographystyle{plain}
\nocite{*}
\bibliography{References}

\appendix

\section{Appendix}
\label{appendix}

\subsection{Method Details}
\subsubsection{Autoencoder}
The autoencoder maps each atom of the input molecular graph into a lower-dimensional latent space, performs diffusion there, and subsequently decodes it to reconstruct the molecule. In this work, we utilize a simple autoencoder instead of a variational autoencoder, which is the commonly adopted approach in related literature~\cite{xu2023geometric}.
 
The encoder and decoder consist of 2 layers of PNA. On the decoder side, the PNA decoder takes these latent embeddings and generates a set of features for each node, referred to here as the output features. The output features are 4-dimensional. 

The edge predictor layer then takes these output features and uses them to determine the presence or absence of edges between node pairs. The edge predictor layer consists of a fully connected layer with ReLU activation to predict the existence of bonds in a molecule. We compare the output of the edge predictor layer with the input graph to compute the MSE loss function and train the model. 

To predict the type of bonds in a molecule, we propose a novel architecture that separates bond type prediction from edge prediction, resulting in more accurate molecule generation. The edge-type prediction model takes the generated molecular graph as input and predicts the bond types for edges in the generated molecule. The edge-type prediction model consists of two graph convolutional layers with ReLU activation followed by a fully connected layer to predict the types of edges in a molecule.

While the graph autoencoder encodes information about the graph structure, the atom-type autoencoder creates embeddings for the atom-type information of each node. The input to this autoencoder is the atom type of all nodes in a molecule, provided in a one-hot encoded format. The encoder converts this input into 2-dimensional atom-type embeddings. The decoder then takes these embeddings and reconstructs the one-hot encoded atom types as output. In latent space, the atom-type embeddings are concatenated with the graph embeddings, and diffusion is applied to the combined set of embeddings.

In the input space diffusion approach, we performed diffusion directly in the input space of the graph rather than converting it to latent space. We use an autoencoder to compress the input node features to the desired dimensions for diffusion. As a first step, we construct a graph from the input graph, modeling the edges as nodes. For example, if the original graph has three nodes (1, 2, 3), the updated graph includes new nodes (4, 5, 6) for each pair of nodes in the original graph (1, 2, 3). The node features of each new node indicate the presence or absence of edges between the corresponding node pairs in the original graph. For example, the node features of the newly added node four could contain information about the presence or absence of edges between nodes one and two. We also construct an edge index for the new graph. We feed this updated graph, which includes edges as nodes along with the edge index, to the encoder to produce node embeddings and perform diffusion. The decoder then takes these node embeddings and reconstructs the original graph. We use a 4-layer graph convolutional network model as both the encoder and decoder, as it replicates the performance of PNA-based encoders and decoders while offering a simpler architecture and reduced training time.

\subsubsection{Gaussian Diffusion}
Standard Gaussian diffusion uses Gaussian noise for the degradation process. The degradation process applies a linear beta schedule, with start and end values of 0.0001 and 0.02, respectively. We define the linear beta schedule to determine how the noise variance \(\beta_t\) increases linearly over time. We also define the corresponding \(\alpha_t\) and its cumulative product \(\bar{\alpha}_t\) as \(\alpha_t = 1 - \beta_t\) and \(\bar{\alpha}_t = \prod_{i=1}^t \alpha_i\), respectively. The forward diffusion process is expressed as \(q(\mathbf{x}_t | \mathbf{x}_0) = \mathcal{N}(\mathbf{x}_t; \sqrt{\bar{\alpha}_t} \mathbf{x}_0, (1 - \bar{\alpha}_t)\mathbf{I})\), where we sample the value for \(\mathbf{x}_t\) as \(\mathbf{x}_t = \sqrt{\bar{\alpha}_t} \mathbf{x}_0 + \sqrt{1 - \bar{\alpha}_t} \mathbf{\epsilon}\), with \(\mathbf{\epsilon} \sim \mathcal{N}(0, \mathbf{I})\).

We use a 7-layer GNN model to restore the latent features. We use 50 steps to degrade the embeddings during training. At each training step, we select a random timestep and degrade the combined embeddings according to the timestep information. The restoration process begins with the noisy data \(\mathbf{x}_t\) and its associated timestep embedding \(h_t\). We concatenate these inputs to form the feature vector \(\mathbf{s}_t = \text{concat}(\mathbf{x}_t, h_t)\), which we pass through the GNN \(f_\theta\). We create a complete graph-based edge index for the embeddings and input it to the GNN/EGNN model to restore the embeddings. From the GNN output, we compute the predicted noise \(\mathbf{z}_t\) as \(\mathbf{z}_t = \mathbf{d}_t - \mathbf{s}_t\). We compare this predicted information loss with the actual information loss to generate an MSE loss signal, guiding the model’s training.

\paragraph{De Novo Generation.}

The generation process begins with fully noised embeddings, sampled from a Gaussian prior distribution \(\mathbf{x}_T \sim \mathcal{N}(\mathbf{0}, \mathbf{I})\); these represent the latent representation at the final timestep \(T\). The model iteratively refines this noisy state, proceeding from \(t = T\) to \(t = 0\). At each timestep \(t\), we concatenate the current data state \(\mathbf{x}_t\) with a normalized timestep embedding \(h_t = \frac{t}{T}\), which encodes time information. We then pass the resulting combined input to the restoration model.

The restoration model predicts the noise added at that stage, \(\mathbf{z}_t\). Using this prediction, we calculate the posterior mean, \(\mu_t\), as~\cite{ho2020denoising}:
\[
\mu_t = \sqrt{\frac{1}{\alpha_t}} \left( \mathbf{x}_t - \frac{\beta_t \mathbf{z}_t}{\sqrt{1 - \bar{\alpha}_t}} \right),
\]
where \(\alpha_t = 1 - \beta_t\), \(\beta_t\) represents the noise variance for the current timestep, and \(\bar{\alpha}_t = \prod_{i=1}^t \alpha_i\) is the cumulative product of \(\alpha_t\) over all timesteps. The posterior mean \(\mu_t\) provides the estimate of the denoised data at timestep \(t-1\).

We compute the posterior variance \(\sigma_t^2\) as:
\[
\sigma_t^2 = \frac{\beta_t (1 - \bar{\alpha}_{t-1})}{1 - \bar{\alpha}_t}.
\]
We sample the denoised data for the previous timestep as:
\[
\mathbf{x}_{t-1} = \mu_t + \sigma_t \mathbf{\epsilon}, \quad \mathbf{\epsilon} \sim \mathcal{N}(0, \mathbf{I}).
\]
After the final timestep \(t = 0\), we give the restored latent representation to the decoders of the autoencoder and atom autoencoder models, as well as the edge predictor model, to generate the molecule graph.

\subsubsection{Heat Dissipation}
\label{subsec:heat_dissipation}

In the heat dissipation approach, the heat equation blurs the 1D embeddings after the encoder stage. We first convert the embeddings to the frequency domain using the Discrete Cosine Transform (DCT), represented as \(\mathbf{X} = \text{DCT}(\mathbf{x})\), where \(\mathbf{X}\) contains the DCT coefficients of the embeddings \(\mathbf{x}\). In the frequency domain, the DCT coefficients are attenuated using a decaying exponential function dependent on the blur's standard deviation \(\sigma\): \(\mathbf{X}' = \mathbf{X} \cdot \exp\left(-\text{freqs}^2 \cdot \frac{\sigma^2}{2}\right)\), where \(\text{freqs} = \frac{\pi}{\text{input\_size}} \cdot [0, 1, 2, \dots, \text{input\_size}-1]\) represents the normalized frequencies. This operation simulates a blur by attenuating higher frequencies. We then transform the blurred embeddings back into the spatial domain using the Inverse Discrete Cosine Transform (IDCT): \(\mathbf{x}' = \text{IDCT}(\mathbf{X}')\).

We add Gaussian noise with a variance of \(\sigma^2\) to the blurred embeddings: \(\mathbf{x}_{\text{noisy}} = \mathbf{x}' + \mathbf{\epsilon}\), where \(\mathbf{\epsilon} \sim \mathcal{N}(0, \sigma^2)\). We train a GNN model to reverse this heat equation-based blurring by predicting the difference needed to restore the embeddings of the previous timestep. The GNN takes in the noisy embeddings \(\mathbf{x}_{\text{noisy}}\) and a complete graph edge index and predicts the difference \(\Delta\mathbf{x} = f_\theta(\mathbf{x}_{\text{noisy}}, \text{edge\_index})\), where \(f_\theta\) is the GNN. We compute the restored embeddings as \(\mathbf{x}_{\text{restored}} = \mathbf{x}_{\text{noisy}} + \Delta\mathbf{x}\). During training, the model minimizes a Mean Squared Error loss by comparing the restored embeddings \(\mathbf{x}_{\text{restored}}\) with the actual embeddings from the previous timestep: \(\mathcal{L} = \text{MSE}(\mathbf{x}_{\text{less\_blurred}}, \mathbf{x}_{\text{restored}})\). Additionally, since the parameters of the blurring framework are fixed, we added an additional KL Divergence loss term to the standard MSE loss term.

\paragraph{De Novo Generation.}

The generation process begins with the encoding of node features using a pre-trained autoencoder. The encoder maps the input features into a latent space representation, which is transformed using an exponential function to ensure positive values. We then blur the resulting embeddings using a heat dissipation mechanism described in Section~\ref{subsec:heat_dissipation}. The sample corresponding to a high level of blur is selected as the starting point for the deblurring process.

The iterative deblurring phase begins with the blurred embeddings. At each step \(i\), we generate a fully connected graph with self-loops to capture the graph structure, and the current embeddings are passed through a deblurring model. This model predicts the mean \(\mathbf{u}_{\text{mean}}\) for the previous timestep. Gaussian noise $\mathbf{\epsilon} \sim \mathcal{N}(0, \mathbf{I})$ is added to the mean, and the embeddings are updated as $\mathbf{u}_{i-1} = \mathbf{u}_{\text{mean}} + \mathbf{\epsilon} \cdot \eta$, where $\eta$ is a small scaling factor that introduces controlled stochasticity. This iterative process removes the blur from the embeddings, progressing toward the unblurred state. After completing the deblurring steps, we give the generated embeddings to the decoders of the autoencoder and atom autoencoder models, as well as the edge predictor model, to generate the molecule graph.

\subsubsection{Flow Matching}

In the flow matching approach, the architecture leverages Optimal Transport conditional vector fields to align source and target distributions. The Optimal Transport framework identifies the most efficient way to transform one probability distribution into another by minimizing a transport cost. This principle is applied in the flow-matching architecture to define continuous flows that interpolate between the source and target distributions.

Firstly, the framework constructs a velocity field to represent the instantaneous velocity required to transform the source distribution into the target distribution. We define it as \(d_\psi = \mathbf{x}_1 - (1 - \sigma_{\min}) \mathbf{x}_0\), where \(\mathbf{x}_0\) is the source embedding, \(\mathbf{x}_1\) is the target embedding, and \(\sigma_{\min}\) is a scaling parameter to ensure numerical stability.

Next, we define the interpolation function, referred to as the conditional flow \(\psi_t\), to describe the evolution of points between the source and target embeddings at any time \(t\). It is defined as \(\psi_t(\mathbf{x}_0, \mathbf{x}_1, t) = \left(1 - (1 - \sigma_{\min}) t\right) \mathbf{x}_0 + t \mathbf{x}_1\). This function smoothly transitions between \(\mathbf{x}_0\) and \(\mathbf{x}_1\) over time, capturing the trajectory of embeddings during the transformation. The interpolation function provides the path or trajectory for the transformation, while the velocity field \(d_\psi\) determines the instantaneous velocity required to move a sample along this trajectory at any given time \(t\).

A GNN-based model predicts the vector field \(v_\theta\). The GNN processes node embeddings over a dynamically created complete graph structure. It comprises an initial graph convolutional layer, 10 hidden layers with ReLU activations, and an output layer. To capture temporal dynamics, sinusoidal time encoding is incorporated into the model. Additionally, the GNN is augmented with a function that generates a complete graph edge index based on the number of latent space embeddings, which is used as input to the graph convolutional layers.

The flow-matching loss aligns the predicted vector field \(v_\theta\) with the constructed velocity field \(d_\psi\). The loss is defined as \(\mathcal{L}_{\text{flow}} = \mathbb{E}_{t, \mathbf{x}_0, \mathbf{x}_1} \left\|v_\theta(t, \psi_t(\mathbf{x}_0, \mathbf{x}_1, t)) - d_\psi \right\|^2\), where \(\mathbf{x}_0\) represents Gaussian noise sampled embeddings, and \(\mathbf{x}_1\) represents the target embeddings.

This framework captures the smooth evolution of embeddings over time, enabling forward and backward integration between source and target distributions. We perform continuous-time integration using an ODE solver. Forward integration maps embeddings from the base distribution at \(t=0\) to the target distribution at \(t=1\), while backward integration reverses the process.

We implement the ODE solver using the \texttt{odeint} function from the zuko library, which numerically approximates these integrals over specified time intervals. We perform forward and backward integration using the \texttt{encode} and \texttt{decode} functions, which map embeddings from \(t=1\) to \(t=0\) and vice versa. The \texttt{encode} function performs backward integration to transform embeddings from the target time \(t=1\) to the base time \(t=0\), while the \texttt{decode} function performs forward integration in the opposite direction. The training objective focuses on aligning the flow predicted by the model with the true transformation of the data.

\paragraph{De Novo Generation.}
The generation process begins by sampling the embeddings from a standard normal distribution \(\mathcal{N}(0, I)\), representing the base distribution at \(t=0\).

The decoding process uses the model to map the sampled embeddings from the base distribution at \(t=0\) to the target distribution at \(t=1\). The resulting embeddings \(x_1\) represent the generated embeddings and are given to the decoders of the autoencoder and atom autoencoder models, as well as the edge predictor model, to generate the molecule graph.

\subsection{Edge-Type Prediction}
The edge-type prediction model takes the molecular graph as input and predicts the bond types in the molecular graph using a GNN architecture. The model takes as input the node feature and edge index matrix. The GNN processes the node features, producing updated embeddings for each node by aggregating information from its neighbors. To predict bond types, the model constructs an edge-level representation for each edge by concatenating the updated embeddings of its two connected nodes. Specifically, for an edge $(i, j)$ in the graph, the edge representation is $[x_i \, || \, x_j]$, where $x_i$ and $x_j$ are the embeddings of nodes $i$ and $j$, respectively, and $||$ denotes concatenation. These edge representations are processed to produce output logits corresponding to bond types. We train the model using a cross-entropy loss, minimizing the difference between the predicted bond types and the ground truth labels.

\subsection{Experimental Details}

We trained the model in the Colab environment with an Intel Xeon CPU @ 2.20GHz, featuring 2 logical processors and 13 GB of RAM. We loaded the QM9 dataset and converted the SMILES to graphs using the RDKit library. We implemented the autoencoder model with the help of the Python torch geometric library. We trained all models for 20 epochs using an Adam optimizer with a learning rate of 0.001.

We use the MSE loss function to train the autoencoder and diffusion models. For the PNA encoder, we use mean, minimum, maximum, and standard deviation aggregators. We can use up to four layers of PNA for encoding to increase accuracy. Increasing the number of layers in the encoder also produces embeddings that are very high in value and have high variance. Having embeddings with high variance can be disadvantageous while training diffusion models; hence, we used only two layers of the PNA encoder and decoder while training the autoencoder. The accuracy achieved by the 2-layer PNA autoencoder was nearly identical to that of the four-layer PNA autoencoder.

The edge classifier in our model uses a neural network architecture. It consists of two fully connected layers, with hidden dimensions of 32 and a ReLU activation function to introduce non-linearity. The edge-type prediction model comprises two graph convolutional layers and a multi-layer perceptron for edge type classification.

For the 1D Heat Dissipation model, the mean and variance of the KL loss term incorporated were 20 and 4, respectively. We calculated the validity of the molecules using the RDKit library. We consolidate the results in Table~\ref{tab:results_table}.

\begin{table}[h!]
\centering
\caption{Summary of experimental results comparing different models, highlighting their validity, uniqueness, and novelty, as well as the training times for the autoencoder and diffusion models given in sec.}

\label{tab:results_table}
\resizebox{\textwidth}{!}{ 
\begin{tabular}{llcccccc}
    \toprule
    & & \multicolumn{5}{c}{\textbf{Model}} \\
    \cmidrule(lr){3-7}
    \textbf{Latent} & \textbf{Metric} & \textbf{GNN Diffusion} & \textbf{EGNN Diffusion} & \textbf{Graph Diffusion} & \textbf{Heat Diffusion} & \textbf{Flow Matching} \\
    \textbf{Space} & & & & & & \\
    \midrule
    \multirow{5}{*}{1D} 
    & Validity & - & - & - & 90\% & - \\
    & Uniqueness & - & - & - & 16\% & - \\
    & Novelty & - & - & - & \textbf{98\%} & - \\
    & Autoencoder- & - & - & - & 2563.2 & - \\
    & Training Time & & & & & \\
    & Diffusion- & - & - & - & 2840.6 & - \\
    & Training Time & & & & & \\
    \midrule
    \multirow{5}{*}{2D} 
    & Validity & \textbf{92\%} & 41\% & 41\% & - & 86\% \\
    & Uniqueness & 28\% & 21\% & 39\% & - & 30\% \\
    & Novelty & 79\% & 77\% & 87\% & - & 71\% \\
    & Autoencoder- & \textbf{3108.8} & 4180.8 & 4609.6 & - & 3108.8 \\
    & Training Time & & & & & \\
    & Diffusion- & 3115.2 & 5654.8 & 8576.7 & - & \textbf{2487.4}
     \\
    & Training Time & & & & & \\
    \midrule
    \multirow{5}{*}{6D} 
    & Validity & 87\% & 34\% & 33\% & - & 68\% \\
    & Uniqueness & 25\% & 30\% & \textbf{40\%} & - & 32\% \\
    & Novelty & 75\% & 88\% & 66\% & - & 81\% \\
    & Autoencoder- & 3524.2 & 4958.2 & 4864.7 & - & 3524.2 \\
    & Training Time & & & & & \\
    & Diffusion- & 3082.6 & 6807.3 & 8302.2 & - & 2604.8 \\
    & Training Time & & & & & \\
    \bottomrule
\end{tabular}
} 
\end{table}

\begin{figure}[h!]
    \centering
    \includegraphics[width=0.8\textwidth]{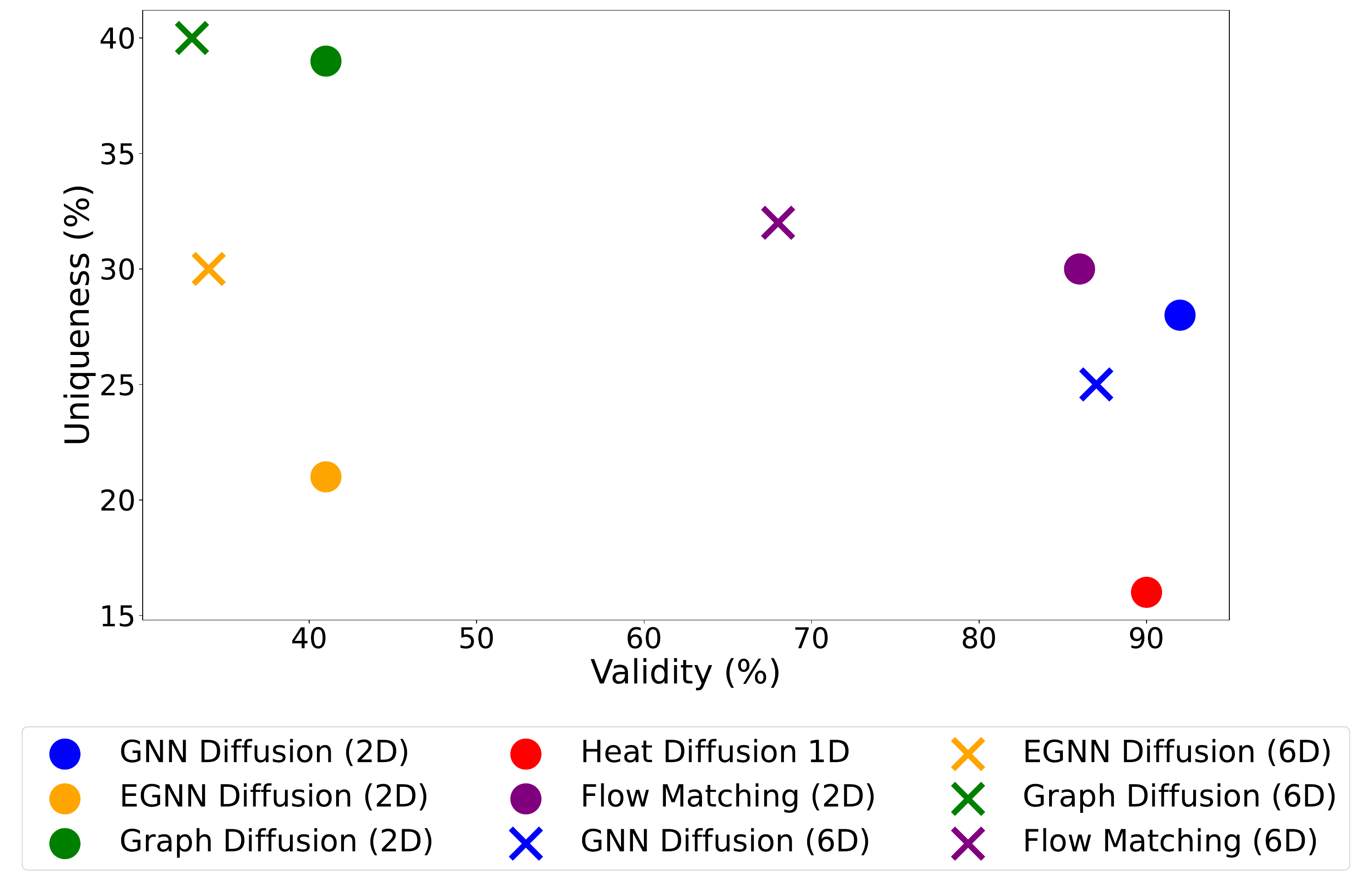}
    \caption{This scatter plot shows the relationship between validity and uniqueness across different model configurations.}

    \label{fig:validity_vs_uniqueness}
\end{figure}

\begin{figure}[h!]
    \centering
    \includegraphics[width=0.8\textwidth]{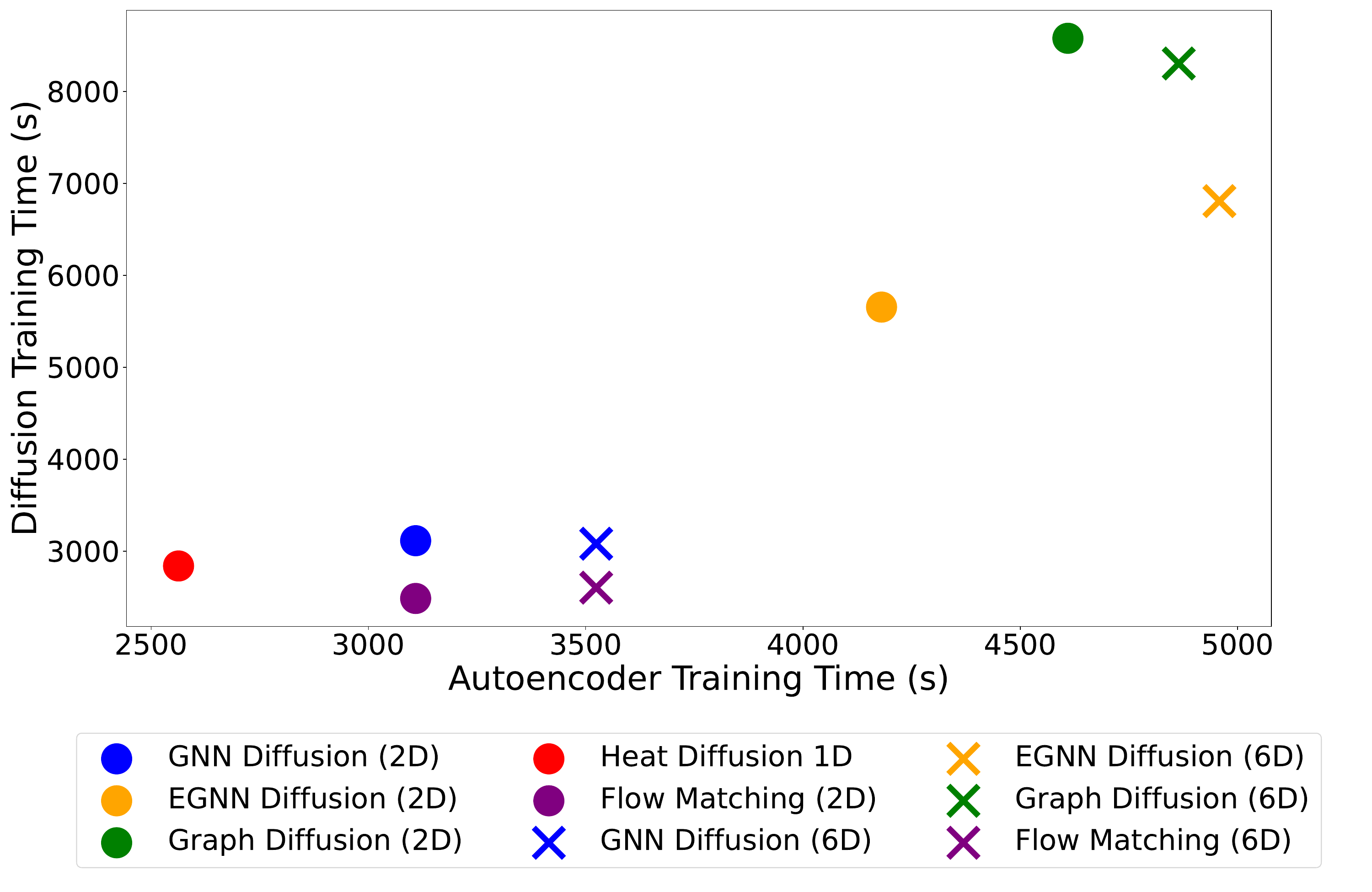}
    \caption{This scatter plot illustrates the relationship between autoencoder training time and diffusion model training time across different model configurations.}

    \label{fig:training_time}
\end{figure}

\begin{figure}[h!]
    \centering
    \includegraphics[width=0.8\textwidth]{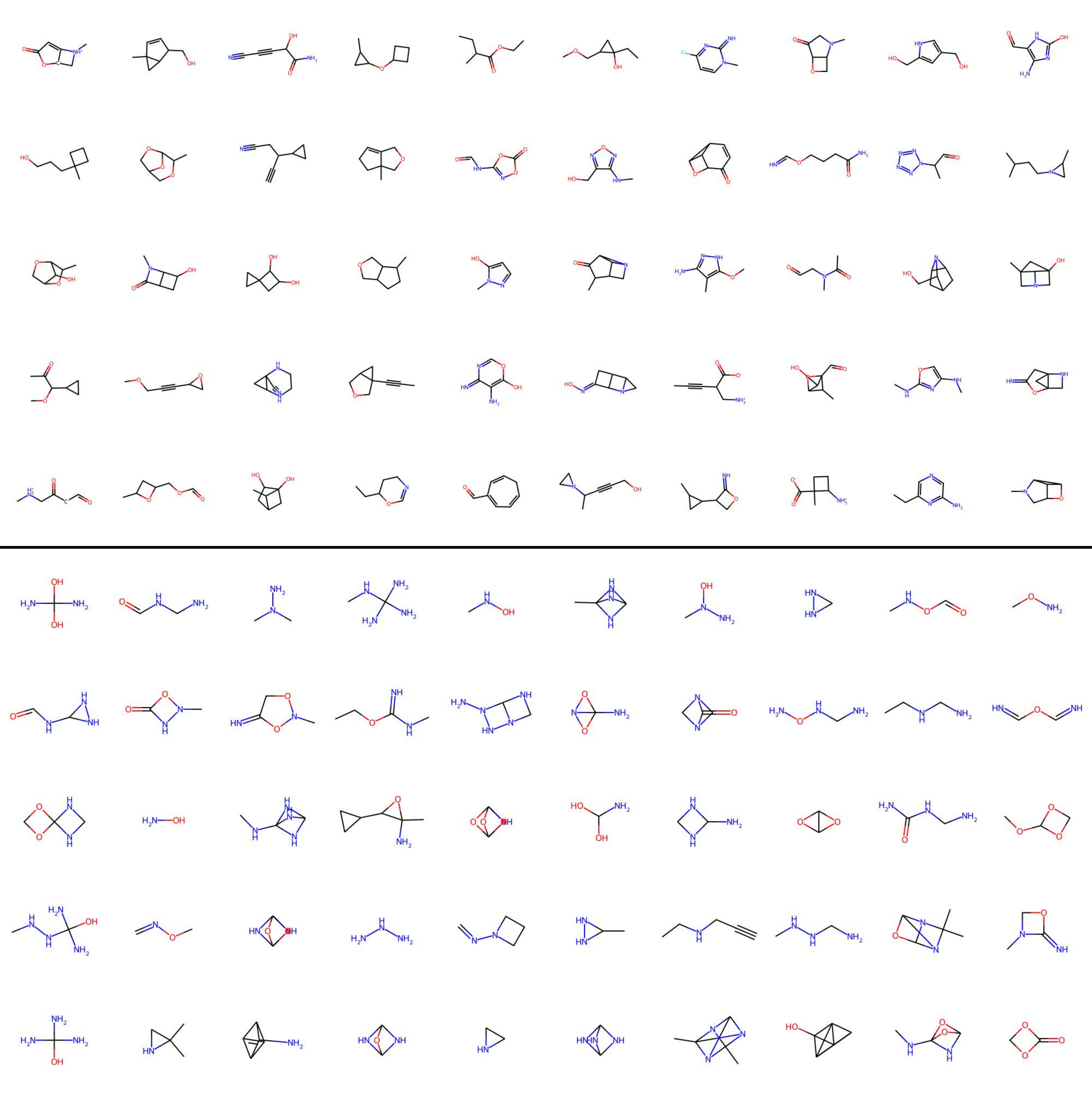}
    \caption{Comparison of molecular structures from the QM9 dataset (top) with generated molecules (bottom). The top half displays 50 randomly selected molecules from the QM9 dataset, while the bottom half showcases 50 randomly selected valid, unique, and novel molecules generated using a GNN-based diffusion model.}

    \label{fig:generated_molecules}
\end{figure}

A scatter plot between the validity and uniqueness of the different models is given in Figure \ref{fig:validity_vs_uniqueness} and a scatter plot between the training times of the autoencoder and diffusion model across different models is given in Figure \ref{fig:training_time}. A comparison of molecular structures from the QM9 dataset with generated molecules is shown in Figure~\ref{fig:generated_molecules}.

\end{document}